\documentclass[10pt,letterpaper]{article}

\usepackage{ccn}
\usepackage{pslatex}
\usepackage{apacite}
\usepackage{graphicx}
\usepackage{booktabs}
\usepackage[implicit=false]{hyperref}

\title{Decoding fMRI Data into Captions using Prefix Language Modeling}

\author{{\large \bf Vyacheslav Shen  (shen9910@kaist.ac.kr)} \\
   School of Electrical Engineering, KAIST\\
291 Daehak-ro, Yuseong-gu, Daejeon, 34141 South Korea
  \AND 
 {\large \bf Kassymzhomart Kunanbayev (kkassymzhomart@kaist.ac.kr)} \\
  School of Electrical Engineering, KAIST\\
291 Daehak-ro, Yuseong-gu, Daejeon, 34141 South Korea
  \AND {\large \bf Dae-Shik Kim (daeshik@kaist.ac.kr)} \\
  School of Electrical Engineering, KAIST\\
291 Daehak-ro, Yuseong-gu, Daejeon, 34141 South Korea
}

\begin{document}

\maketitle

\clearpage

\section{Abstract}
{
\bf

With the advancements in Large Language and Latent Diffusion models, brain decoding has achieved remarkable results in recent years. The works on the NSD dataset, with stimuli images from the COCO dataset, leverage the embeddings from the CLIP model for image reconstruction and GIT for captioning. However, the current captioning approach introduces the challenge of potential data contamination given that the GIT model was trained on the COCO dataset.  In this work, we present an alternative method for decoding brain signals into image captions by predicting a DINOv2 model's embedding of an image from the corresponding fMRI signal and then providing its [CLS] token as the prefix to the GPT-2 language model which decreases computational requirements considerably. Additionally, instead of commonly used Linear Regression, we explore 3D Convolutional Neural Network mapping of fMRI signals to image embedding space for better accounting positional information of voxels.

The code of our work is available by the following link:\hfill\url{https://github.com/slavaheroes/brain-captioning-with-gpt2} 

}
\begin{quote}
\small
\textbf{Keywords:} 
brain decoding; brain captioning; fMRI;
\end{quote}

\section{Introduction}

It has been a common practice for brain decoding studies on the Natural Scenes Dataset (NSD) \cite{Allen2021AM7} to predict the embeddings of multimodal vision-language models like CLIP \cite{radford2021learning}, GIT \cite{wang2022git} from brain activations to use these predictions for further image and caption generation. These approaches typically involve mapping high-dimensional fMRI data (e.g., 15,724 voxels for subject 1) to even higher-dimensional model embeddings. Current state-of-the-art brain captioning methods \cite{ferrante2023brain} \cite{scotti2024mindeye2}, which primarily leverage the GIT model (with an embedding dimension of $257\times1024$) for caption generation from fMRI signals, require considerable computational resources due to these high-dimensional transformations. Also, in the conventional NSD train/test split \cite{Takagi_2023_CVPR} \cite{ozcelik2023natural}, all images from the test set appear in the training set of GIT, raising the concern of possible data contamination.

Additionally, in other brain decoding works including \cite{mai2023unibrain} \cite{ozcelik2023natural} \cite{Takagi_2023_CVPR}, fMRI voxels for a particular image are linearized using an ROI mask, followed by the application of Ridge Regression to map the fMRI voxels to the model embeddings. However, activations from regions not present in the ROI mask could be overlooked, and positional information of these voxels, which might help in brain decoding, is neglected. 

To address the mentioned challenges, our study tests a new method utilizing a single DINOv2 \cite{oquab2023dinov2} [CLS] vector (of size $1536$)  as a representation of an image and the GPT-2 language model \cite{radford2019language} for text generation. The main idea is to predict the DINOv2 embedding directly from the corresponding fMRI signals using 3D-Convolutional ResNets \cite{he2016deep}, and pass them as a prefix through the language model to generate captions as proposed in \cite{mokady2021clipcap}.

\section{Methodology}

\subsection{Dataset preprocessing}

We follow the traditional NSD train/test split and data preprocessing of 4 subjects (sub1, sub2, sub5, sub7) obtained from GLM (\textit{betas\_fithrf\_GLMdenoise\_RR}) as it was utilized in Ozcelik and VanRullen \citeyear{ozcelik2023natural}.

Given that, fMRI data represents a 4D array ($time \times W \times D \times H$), for Ridge Regression mapping from brain activity to the DINOv2 embedding space. We applied z-normalization for the linearized fMRI voxels\footnote{For sub1, input is a 1D vector of shape 15724} extracted from the NSDGeneral ROI mask, whereas for CNNs, a 3D input\footnote{For sub1, input has (81, 104, 83) shape} at each time point was scaled between -1 and 1.

\subsection{Brain Captioning}

\begin{figure*}[h]
\centering
\includegraphics[width=0.9\textwidth]{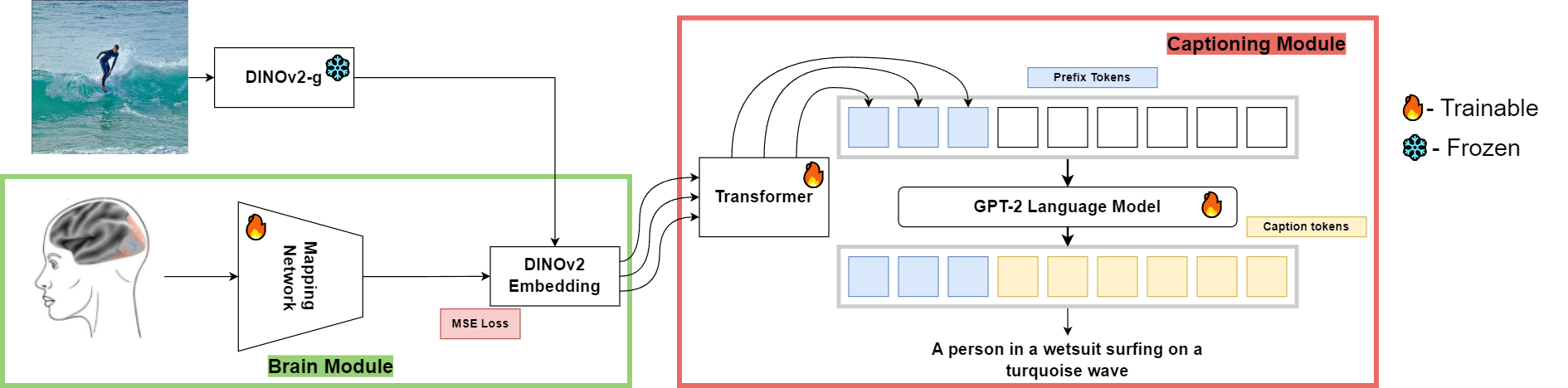}
\caption{The scheme of our method. GPT-2 base model was used as the language model, while fMRI-DINOv2 embedding mapping } 
\label{fig1}
\end{figure*}

The scheme of our approach is presented in Figure \ref{fig1}. It has two parts: a brain and a captioning module. Both modules are trained separately.

The brain module is used to map fMRI activations into a DINOv2 embedding. DINOv2 was chosen due to its rich and robust visual features achieved by self-supervised learning compared to CLIP. It is trained using Mean Squared Error (MSE) loss where the ground truth label is the image embedding from the DINOv2-g model. We tried 3 mapping networks: Ridge Regression and two variations of 3D ResNet with 18 layers referred to as Shallow and Wide CNNs. Wide CNN has more feature planes than Shallow CNN.

The captioning module consists of a light transformer \cite{vaswani2017attention} and a language model. During training, the transformer converts the DINOv2 image embedding into prefix tokens that have the same dimensions as a word embedding. These prefix tokens are then used as inputs for the language model. The training objective is to predict caption tokens conditioned on the prefix autoregressively \cite{mokady2021clipcap}.

At inference time, the brain module predicts the embedding of the seen image from the fMRI signal, which is then passed to the captioning module. While decoding from the language model, the beam search was employed to select the next token.

The training hyperparameters and design choices can be found in the provided GitHub link.

We compare the effectiveness of our approach with Ferrante et al. \citeyear{ferrante2023brain}, UniBrain \cite{mai2023unibrain}, MindEye-2 \cite{scotti2024mindeye2}.

\section{Results}

\begin{table*}[!h]
\centering
\begin{center}
\caption{Evaluation of captions from fMRI for sub1} 
\label{tab1}
\begin{tabular}{p{0.1\linewidth}ccc|ccc}
\cline{2-7}
\multicolumn{1}{l}{}        & \multicolumn{3}{c|}{fMRI vs COCO}                                                        & \multicolumn{3}{c}{fMRI vs Model}                                             \\ 
\midrule
\multicolumn{1}{l}{Metrics} & \multicolumn{1}{l}{UniBrain} & \multicolumn{1}{l}{MindEye-2} & \multicolumn{1}{l|}{Ours} & \multicolumn{1}{l}{Ferrante et al. \citeyear{ferrante2023brain} } & \multicolumn{1}{l}{MindEye-2} & \multicolumn{1}{l}{Ours} \\ 
\midrule
METEOR                       & 0.170                        & 0.248                         & \textbf{0.271}            & 0.305                        & 0.344                         & \textbf{0.457}           \\
ROUGE-1                      & 0.247                        & \textbf{0.353}                & 0.346                     & -                            & 0.455                         & \textbf{0.513}           \\
ROUGE-L                      & 0.225                        & \textbf{0.326}                & 0.316                     & -                            & 0.427                         & \textbf{0.491}           \\
Sentence                     & -                            & \textbf{47.9\%}                & 39.7\%                     & 44.7\%                        & \textbf{52.3\%}                & 49.9\%                    \\
CLIP-B                       & -                            & \textbf{73.7\%}                & 68.7\%                     & 70.5\%                        & \textbf{75.4\%}                & 74.6\%                    \\
CLIP-L                       & \textbf{86.1\%}               & 63.8\%                         & 58.5\%                     & -                            & 67.1\%                         & \textbf{67.4\%}          \\
\bottomrule
\end{tabular}
\end{center}
\end{table*}

Table \ref{tab1} presents our Wide CNN results in comparison with existing works. Following the evaluation protocol used by Scotti et al. \citeyear{scotti2024mindeye2}, we assess captions predicted by our model with two different ground truth captions: the original COCO captions and captions generated directly by our captioning module. The latter shows how well our approach predicts the behavior of our captioning module,

When evaluated against the original COCO captions, our approach demonstrates superior performance in the METEOR metric and achieves comparable ROUGE scores to MindEye-2, while requiring only 1/171th of the parameter space. More significantly, the right half of Table \ref{tab1} shows that our method outperforms in 4 metrics out of 6 when comparing against captions generated from true image embeddings. It indicates both the efficiency of the brain module and more accurate representational mapping of DINOv2 image embeddings from fMRI data.

\begin{table}[!ht]
\caption{Evaluation of Mapping Networks from fMRI to DINOv2 embedding space averaged for 4 subjects} 
\label{tab2}
\begin{tabular}{p{0.27\linewidth}ccc}
\cline{2-4}
         & \multicolumn{3}{c}{fMRI vs COCO}                                                           \\ \hline
Metrics  & \multicolumn{1}{l}{Ridge} & \multicolumn{1}{l}{Shallow CNN} & \multicolumn{1}{l}{Wide CNN} \\ \hline
METEOR   & 0.263              & 0.267                  & \textbf{0.273}        \\
ROUGE-1  & 0.331              & 0.340                   & \textbf{0.346}        \\
ROUGE-L  & 0.300             & 0.312                    & \textbf{0.317}        \\
Sentence & 34.92\%            & 36.71\%                 & \textbf{38.91\%}       \\
CLIP-B   & 66.73\%            & 67.22\%                 & \textbf{67.79\%}       \\
CLIP-L   & 55.72\%            & 56.65\%                 & \textbf{57.59\%}      \\
\bottomrule
\end{tabular}
\end{table}
\subsection{Ablation on Mapping Network}

Table \ref{tab2} shows the results of different mapping networks used in the brain module. The advantages of CNN architectures compared to Ridge Regression, particularly our Wide CNN implementation, are evident across all metrics. This superior performance suggests the importance of capturing both out-of-ROI voxel contributions and positional information between voxels, which linear mapping approaches cannot capture.

\section{Conclusion \& Future Work}

This work demonstrates competitive performance in brain captioning while significantly reducing computational requirements compared to previous approaches. Our approach minimizes the issue of data contamination to zero by using the DINOv2 model which has not been trained on the COCO dataset, and introduces a new method to fMRI signals to the image embedding space by passing complete fMRI volume through the convolutional neural networks. This improvement implies that information outside of the ROI mask and positional information can enhance brain decoding.

Also, providing fMRI data as the prefix for the language models \cite{ye2023language} can be promising in creating brain decoding frameworks designed for complex tasks like visual-question answering.

\section{Acknowledgments}
This work was supported by the Engineering Research Center of Excellence (ERC) Program supported by National Research Foundation (NRF), Korean Ministry of Science \& ICT (MSIT) (Grant No. NRF-2017R1A5A101470823).

\bibliographystyle{apacite}

\setlength{\bibleftmargin}{.125in}
\setlength{\bibindent}{-\bibleftmargin}

\bibliography{ccn_style}

\end{document}